\documentclass[10pt,dvipsnames,svgnames]{article} % For LaTeX2e
\usepackage[preprint]{tmlr}

\usepackage{amsmath,amsfonts,bm}

\def\eqref#1{equation~\ref{#1}}
\def\1{\bm{1}}

\DeclareMathAlphabet{\mathsfit}{\encodingdefault}{\sfdefault}{m}{sl}
\SetMathAlphabet{\mathsfit}{bold}{\encodingdefault}{\sfdefault}{bx}{n}

\usepackage{xcolor}
\usepackage{graphicx}
\usepackage{subfigure}
\usepackage{booktabs} % for professional tables
\usepackage{multirow}
\usepackage{collcell,eqparbox,makecell}
\usepackage{hyperref}
\usepackage{url}
\usepackage{algorithm}
\usepackage{algorithmic}

\let\classAND\AND
\let\AND\relax
\usepackage{algorithmic}
\let\AND\classAND
\AtBeginEnvironment{algorithmic}{\let\AND\algoAND}

\floatname{algorithm}{Algorithm}

\usepackage{amsmath}
\usepackage{mathtools}
\usepackage{amsthm}
\usepackage{pdfpages}
\usepackage{amssymb}
\theoremstyle{plain}

\theoremstyle{definition}

\theoremstyle{remark}

\newcommand{\Q}{\bold{Q}}
\newcommand{\K}{\bold{K}}
\newcommand{\V}{\bold{V}}

\newcommand{\R}{\bold{R}}

\newcommand{\X}{\bold{X}}

\usepackage[textsize=tiny]{todonotes}

\title{SAMSA: Efficient Transformer for Many Data Modalities}

\author{\name Minh Lenhat \email minhln30@fpt.com \\
      \addr FPT Software AI Center
      \AND
      \name Viet Anh Nguyen \email anhnv117@fpt.com \\
      \addr FPT Software AI Center
      \AND
      \name Khoa Nguyen \email khoant30@fpt.com\\
      \addr FPT Software AI Center
      \AND
      \name Hieu Duong Duc \email hieudd12@fpt.com\\
      \addr FPT Software AI Center
      \AND
      \name Hung Dao Huu \email hungdh3@fpt.com\\
      \addr FPT Software AI Center
      \AND
      \name Truong Son Hy$^\dagger$ \email thy@uab.edu\\
      \addr University of Alabama at Birmingham \\ FPT Software AI Center }

\def\month{MM}  % Insert correct month for camera-ready version
\def\year{YYYY} % Insert correct year for camera-ready version
\def\openreview{\url{https://openreview.net/forum?id=XXXX}} % Insert correct link to OpenReview for camera-ready version

\begin{document}

% \footnote{Automatically generated footnote markers work fine!}
\maketitle

% \author{\name Minh Lenhat$^*$ \email minhln30@fpt.com \\
%       \addr FPT Software\\
%       \AND
%       \name Viet Anh Nguyen$^*$ \email anhnv117@fpt.com \\
%       \addr FPT Software
%       \AND
%       \name Khoa Nguyen \email khoant30@fpt.com\\
%       \addr FPT Software \\
%       \AND
%       \name Hieu D. D. \email hieudd12@fpt.com\\
%       \addr FPT Software \\
%       \AND
%       \name Hung D. H. \email hungdh3@fpt.com\\
%       \addr FPT Software \\
%       \AND
%       \name Hy Truong Son^$^{**}$ \email TruongSon.Hy@indstate.edu\\
%       \addr Indiana State University \\}
% \def\thefootnote{*}\footnotetext{These authors contributed equally to this work}\def\thefootnote{\arabic{footnote}}
\def\thefootnote{$\dagger$}\footnotetext{Correspondence to thy@uab.edu}\def\thefootnote{\arabic{footnote}}
\begin{abstract}
The versatility of self-attention mechanism earned transformers great success in almost all data modalities, with limitations on the quadratic complexity and difficulty of training. Efficient transformers, on the other hand, often rely on clever data-modality-dependent construction to get over the quadratic complexity of transformers. This greatly hinders their applications on different data modalities, which is one of the pillars of contemporary foundational modeling. In this paper, we lay the groundwork for efficient foundational modeling by proposing \textbf{SAMSA} - SAMpling-Self-Attention, a context-aware linear complexity self-attention mechanism that works well on multiple data modalities. Our mechanism is based on a differentiable sampling without replacement method we discovered. This enables the self-attention module to attend to the most important token set, where the importance is defined by data. Moreover, as differentiability is not needed in inference, the sparse formulation of our method costs little time overhead, further lowering computational costs. In short, SAMSA achieved competitive or even SOTA results on many benchmarks, while being faster in inference, compared to other very specialized models. Against full self-attention, real inference time significantly decreases while performance ranges from negligible degradation to outperformance. We release our source code in the repository: \url{https://github.com/HySonLab/SAMSA}

% We release our source code in the anonymous repository: https://anonymous.4open.science/r/samsa-2E92/.

% It is important to emphasize that our model is a foundation model that can work with multiple types of data structures including point clouds, long-range sequences, and graphs.
\end{abstract}

\section{Introduction} \label{sec:introduction}

Transformers \citep{AttentionAllYouNeed} have been successfully applied to a wide variety of tasks on many data modalities, namely sequence \citep{AttentionAllYouNeed,GPT2,BERT}, vision \citep{VisionTransformer,DETR,LLAVA}, speech \citep{Conformer,AudioSpectrogramTransformer,AudioTransformerPatchout}, time-series \citep{NonStationaryTransformer,TimeLLM}, point cloud \citep{SetTransformer,PointTransformer,GumbelSubsetSampling}, and graph \citep{GraphTransformerNetwork,GraphGPS,Exphormer,10.1063/5.0152833}. This can be attributed to the permutation equivariance of transformers, which provides a unique way of encoding structural information through positional encoding (absolute or relative positional enccodings). Ultimately, transformers become the building blocks of contemporary foundational modelings. While transformer can be the one architecture for all data modalities, quadratic complexity that hinders it from processing very long sequences. To mitigate the problem of quadratic complexity, based on the data modality they are working on (mostly sequence), efficient transformers~\citep{Beltagy2020Longformer,EncodingLongandStructuredInputsinTransformers,sliding_window,BigBird,Exphormer} have been developed. This is great for its original usage and some of the mentioned works have found their way to industrial applications. However, it has a major flaw: it only works for the intended data modality and is often hard to adapt to another data modality. For example, it is unreasonable to expect sliding window attention~\citep{sliding_window} to work on data with non-rigid data structures like point clouds or graphs. Without efficient building blocks working across many data structures, future works toward AGI would be costly in both training and inference.

In this work, we develop towards data modality agnostic efficient transformer architecture to fill the literature gap. Any satisfying architecture should respect the permutation equivariance that full self-attention adheres to for versatility, be both sub-quadratic asymptotic computational cost for very large number of tokens, and be efficient in common cases (i.e. 512 - 1024 tokens) for practical usage. Inspired by the finite-yet-directed nature of human attention, we developed a {parallelizable differentiable sampling without replacement} algorithm that allows models to learn to choose what set of tokens they should attend to. This parallelization of our differentiable sampling algorithm enables our model to be trained efficiently while the without-replacement characteristic allows our model to maintain some functional expressivity. Since the result of one attention head is not dependent on the input of any other attention head, different attention heads can use different sampled sets of tokens. This consequentially enlarges the receptive field, which further increases the expressivity of our model.

To demonstrate the effectiveness of our method on many data modalities, we conduct extensive experiments on multilinear and non-multilinear data modalities. For multilinear data modalities, we experimented on sequence data modality via the long-range arena benchmark \citep{longrangearena} featuring four classification tasks and one retrieval task on these datasets. For non-multilinear data modalities, we experimented on graph and point cloud data modalities: graph classification/regression via peptides dataset \citep{Peptideds} from Long-Range Graph Benchmark \citep{LongRangeGraphBenchmark}, object classification on ModelNet40 dataset \citep{Modelnet40}, and part segmentation on ShapeNet dataset \citep{Shapenet}. Compared to specialized models with data-modality-specific inductive bias, we achieve competitive results while being much faster than them. Against full self-transformers, {the performance degradation was negligible, and in some cases, unexpectedly outperformed full self-attention transformers}. Aside from comparative experiments, we also conducted many ablation studies to understand: the efficiency-performance trade-off, regularization effect of sampling mechanisms, and training stability w.r.t number of sampled tokens and the sparse formulation.

In short, our contributions can be summarized as follows:

\begin{itemize}
    \item Parallelizable differentiable sampling without replacement method and its integration to self-attention mechanism,
    \item Fast training time and negligible extra cost for inference compared to transformers with negligible performance loss,
    \item Demonstrated effectiveness on many data modalities, competitive results even against specialized models.
\end{itemize}

% We provide the detailed theoretical analysis in the Appendix Section~\ref{sec:theory}.
\section{Related works} \label{sec:related}

\paragraph{\textbf{Efficient Transformers.}} The scalability of transformers is hindered by the quadratic complexity of self-attention modules. Hence, numerous works have delved into sparsity \citep{Beltagy2020Longformer,EncodingLongandStructuredInputsinTransformers,BigBird,Exphormer}, low-rank projection \citep{wang2020linformer}, hashing-based method \citep{kitaev2020reformer}, and kernel-based methods \citep{performer,TransformerAreRNN_LinearTransformer} to sacrifice the expressivity of transformers for a sub-quadratic construction. Our review of contemporary efficient transformers is influenced by this survey by \citep{efficienttransformersurvey}. Linformer \citep{wang2020linformer} is based on the low-rank assumption of self-attention matrices, realized via projection matrices. As one of the earliest works in efficient transformers, it has flaws in both scalability and fixed sequence length. Subsequent work like Linear Transformer \citep{TransformerAreRNN_LinearTransformer} redesigns the similarity kernel previously as a softmax of dot-product score (or any other kernels) to one made of a linear probability cage. The modification frees them from the recomputation of the multiplication of key-value matrices, making attention computational cost $O(N)$, where $N$ refers to the number of tokens. Without softmax non-linearity, the rank of those attention matrices is bounded by the number of dimensions of tokens, which loses much expressivity of transformer models. Methods like \citep{Beltagy2020Longformer} utilize three choices of fixed-sparsity-patterns: sliding window, dilated sliding window, and global+sliding window. Similarly, \citep{BigBird} uses a combination of different sparse patterns at one time: random attention, window attention, and global attention. On graphs, \citep{Exphormer} also combines different sparsity patterns but with a constraint on the random-generated one: it has to be an expander graph. The hard-coded sparsity effectively linearized the complexity of transformers and is empirically fast. However, intuitively, no pattern or finite set of patterns fits all problems; therefore, there is also a line of work dedicated to a differentiable sparsity \citep{pmlr-v196-hy22a, pmlr-v228-nguyen24a}. \citep{RoutingTransformer} drew a relationship from MIPS problem (Maximum Inner Product Search) and the Nearest-Neighbor Search algorithm, when both the query and key vectors are unit vectors. With their online k-means algorithm, their method attends each query vector to every key vector belonging to the same cluster, thus, bringing the complexity down to $O(n^{1.5})$, with $n$ is the number of tokens. Practically, however, their method is not fast as it needs a k-means construction and extensive specialized implementation for sparse operators. Another perspective on efficient transformers is making better implementation of vanilla transformers~\citep{MemoryEfficientAttention,Flash1,Flash2}. \citep{MemoryEfficientAttention} proposed a neat trick for attention mechanism: chunking. While the mechanism is simple, they have reduced the memory cost from quadratic to sublinear, effectively lifting the computation constraint of GPU memory. \citep{Flash1} both reduces the frequency of data transfer between HBM and SRAM within the GPU and derives a memory efficient backward pass. This allows quadratic attention scales to the sequence length of 64000 and beyond. \citep{Flash2} is an incremental work from \citep{Flash1} with better work partition and reduction in non-matmul FLOPS.

% \paragraph{\textbf{Expressiveness of Transformers}} Approximation power is an essential theoretical aspect when designing deep learning network, as it aims to ensure the models' learnability given sufficient budget. The vanilla transformer \citep{AttentionAllYouNeed} has been shown to be capable of approximating any sequence-to-sequence function with compact domain, given sufficient depth. Later, sparse Transformer with $O(n)$ connections in the Transformer has been proved to exhibit the same approximation power \citep{SparseTransApprox}. This renovates the motivation of finding transformers variants that are not only efficient but also expressive. However, non-parametric methods like sliding window \citep{sliding_window} or random attention \citep{ReluAndSoftmaxTransformer_relusum} with fixed sparsifying patterns can heavily affect the tokens semantic feature.

\paragraph{\textbf{Continuous Relaxation}} Discrete operations are much used in decision problems (classification, next-word prediction, plan, \dots). However, these operations by nature are not amenable to gradient-based optimization because that operation has gradients everywhere either zero or undefined. For example, the heaviside function for binary choice:
\begin{equation}
H(x) = \begin{cases} 
0 & \text{if } x < 0 \\
1 & \text{if } x \geq 0 
\end{cases}
\end{equation}
has gradient undefined at $x = 0$ and zero everywhere else. Since continuous relaxation of discrete operators and algorithms is a complex study and the sampling operator is the core of our efficient transformer formulation, we focus our literature review on differentiable sampling. To continuously relax the sampling operator, the previous literature was developed based on the relaxation of choose, sort, and top-k algorithms~\citep{GumbelSoftmax,concrete,stochasticbeam, sort_OT, fast_soft_sort, sort_network}. \citep{GumbelSoftmax, concrete} is the foundational work on parameterization of discrete categorical variable. They introduces the Gumbel-Softmax distribution that approximate categorical distribution exceptionally well. There are some differences between the two works: \citep{GumbelSoftmax} introduces straight through estimator via application of path derivative~\citep{stochastic_neurons} and \citep{concrete} takes account of density in their relaxed objective function. From differentiable categorical variable, differentiable sampling with replacement can be obtained trivially by computing these stochastic categorical variable $k$-times, where $k$ is the number of samples. Differentiable sampling without replacement, however, is much trickier and is an active area of research. \citep{stochasticbeam} is one of the earliest work in differentiable sampling without replacement. Their process of sampling without replacement, Gumbel Top-k, samples tokens sequentially with logit normalization, applied in generative text models as stochastic beam search. \citep{sort_OT} sees the differentiable without replacement in another perspective: optimal transport, where loss function represents how far away is the current solution from the optimal one. \citep{fast_soft_sort} is an efficient estimation of sort function differentiation through optimal transport via permutahedron projections. The projection decreases the quadratic computational cost to $O(n\log{}n)$ forward pass and $O(n)$ backward pass. \citep{sort_network} takes a pre-existing discrete parallelizable sorting/top-k algorithm (e.g. odd-even sorting network, bitonic sorting network, selection network, \dots) and relaxes all of the control flows for gradient estimation.

\section{Background} \label{sec:background}
\subsection{Self-Attention Mechanism}
\paragraph{\textbf{Vanilla Self Attention.}} The self-attention mechanism is a core component of Transformer models for long-range dependencies. Let $\mathbf{X}=\left\{\mathbf{x}_1, \mathbf{x}_2, \ldots, \mathbf{x}_n\right\}$ $ \in \mathbb{R}^{n \times d}$ represent a set of tokens of $n$ $d$-dimensional vectors. The matrices $\mathbf{Q}$, $\mathbf{K}$, and $\mathbf{V}$ all in $\R^{n\times d_h}$ representing query, key, and value vectors are created via learnable linear transformation of the input token matrix $\mathbf{X}$: 
\begin{align*}
    \Q &\;= \X W_Q + \mathbf{1}_nb_q^{\top},\\
    \K &\;= \X W_K+ \mathbf{1}_nb_k^{\top},\\
    \V &\;= \X W_V+ \mathbf{1}_nb_v^{\top}.
\end{align*}
Then, using the query, key, and value vectors, it creates another set of tokens by aggregating those value vectors based on the dependencies defined through the attention map constructed by query and value vectors:
\begin{equation} \label{eq:self_attention}
\operatorname{Attention}(\mathbf{X})=\operatorname{Softmax}\left(\frac{\Q\K^{\top}}{\sqrt{d_h}}\right) \mathbf{V},
\end{equation}
This mechanism allows information aggregation of very far tokens and effectively captures complex long-range relationships.

\paragraph{\textbf{Efficient Quadratic Attention.}} Computing the given $\operatorname{Attention}$ function naively yields a quadratic memory cost due to the need to store attention maps for backpropagation. However, its memory complexity can be reduced into linear complexity by checkpointing~\citep{MemoryEfficientAttention} or algebraic manipulation of derivative function~\citep{Flash1}. Via fused kernel~\citep{Flash1,Flash2}, the computational cost of self-attention is further decreased through fewer memory operations between HBM and SRAM. This allows better parallelization, making quadratic attention scales to the length of 64k and beyond.

\subsection{Continuous Relaxation of Top-k Sampling}
\label{sec:relaxation}

\paragraph{\textbf{Top-k Sampling.}} The top-k sampling operation selects the best set of $k$ objects out of others under a criterion, deterministic or stochastic. The criterion function is a permutation invariant function that maps a set of $n$ vectors to a real number. The following equation describes a mathematical formulation of top-k sampling:
\begin{equation}
    \operatorname{Sample}(X, k) = \operatorname{argmax}_S^{C(X, k)}\text{criterion}(S), \label{eq:naive_sampler}
\end{equation}
where $X$ is the set of vector to sample from, $k$ is the number of sampled vectors needing to be sampled, and $C(N,k)$ is the set of all k-combinations of set $N$. Since the $\operatorname{argmax}$ operator has no continuity, the derivative of the sampled vectors with respect to the criterion is either zero or undefined, i.e., not useful. Therefore, it is hard to integrate into neural networks. Early work~\citep{deepknn} had to resort to a two-stage optimization. This circumvention, however, introduces additional complexity and worsens performance via non-aligned objectives.

\paragraph{\textbf{Straight Through Estimator.}} Before going deeper into the differentiable top-k literature, it is crucial to explain a much simpler case: $k=1$. To choose one vector from many, \citep{GumbelSoftmax} provides a differentiable approximation of the categorical distribution via Softmax with Gumbel noise added on logits:
\begin{equation*}
\label{eq:gumbel_softmax}
\operatorname{GumbelSoftmax}(\pi)[i] \triangleq \frac{\exp\left((\pi_i + g_i) / \tau\right)}{\sum_{j=1}^{n} \exp\left((\pi_j + g_j) / \tau\right)},
\end{equation*}
where \( g_i \) are independent and identically samples from \text{Gumbel}(0,1) distribution, \( \pi_i \) are the unnormalized logits, and \( \tau \) is the temperature parameter that controls the smoothness of the distribution. By adjusting $\tau$, the Gumbel-Softmax can transition between a one-hot categorical distribution and a softer distribution, enabling gradient-based optimization. The straight-through estimator, however, uses the above soft-categorical function as a differentiable proxy for the hard-categorical function obtained by $\operatorname{argmax}$:
\begin{align*}
    \operatorname{ST-GumbelSoftmax}(\pi)[i] &\triangleq 
        \begin{cases}
            1,& \text{if } i = \operatorname{argmax}(\pi), \\
            0,              & \text{otherwise},
        \end{cases} \\
    \partial{\operatorname{ST-GumbelSoftmax}} / \partial{\pi} &\triangleq \partial{\operatorname{GumbelSoftmax}} / \partial{\pi}.
\end{align*}
A naive way to extend this to differentiable top-k~\citep{stochasticbeam} is to use Gumbel-Softmax to sample from the unsampled vectors $k$ times sequentially, named Gumbel-Topk. This is not parallelizable and has unreliable gradients; therefore, it cannot be used inside repeating deep learning modules, e.g. transformer layers.

\section{Method} \label{sec:method}
\subsection{Overview}

\begin{figure*}[t]
\begin{center}
\includegraphics[width = 1.0\textwidth]{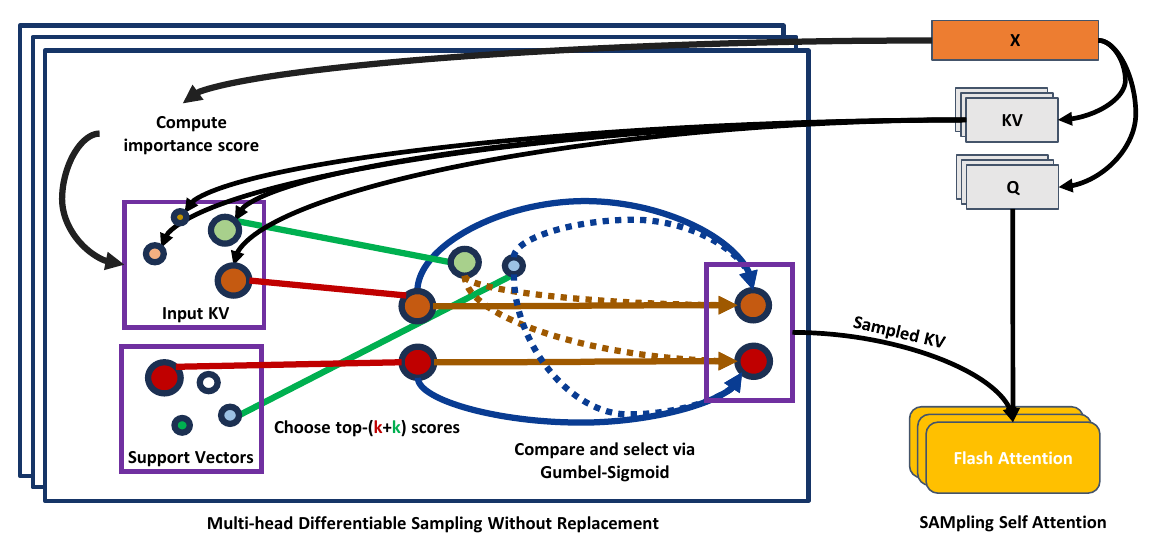}
\caption{Overview of our proposed model sampling-self-attention module SAMSA. The key-value vectors are selected via top-k importance score computed using tokens' latent. The Gumbel-Sigmoid reparameterization provides gradients to guide the optimization process towards most important key-value pairs of vectors (left). The sampled key-value vectors are then fed into Flash Attention to attend to the query vectors (right).}
\label{fig:samsa}
\end{center}
\end{figure*}

Our method is a composition of context-aware sampling without replacement method (Section~\ref{sec:samsa}) and the integration to self-attention mechanism (Section~\ref{sec:intSA}). Via reparameterization via Gumbel-Softmax, we devised a parallelizable differentiable sampling without replacement as an innovation from Gumbel-Top-K in existing literature~\citep{stochasticbeam} - a sequential method. Similar to Linformer \citep{wang2020linformer} applying low-rank approximation to key and value vectors, we attend query vectors to the sampled key and value vectors. Inspired by the MoE architecture where each MLP chooses different subsets of tokens to transform~\citep{expert_choose_token}, our architecture assigns different sets of key-value pairs to different heads. This novel paradigm can be seen as mixture of attention experts, where each expert oversees different sets of tokens. The method is summarized in Algorithm~\ref{alg:sampling_transformer}.

\begin{algorithm}[h]
\caption{Sampling-based Transformer Layer}
\label{alg:sampling_transformer}
\begin{algorithmic}[1]
\REQUIRE Input token matrix $\mathbf{X}$, sample size $k$
\ENSURE Updated token matrix $\mathbf{X}$
\STATE \textbf{Transform tokens to query and key-value matrices (Pre-RMSNorm)}
\STATE $\widehat{\X} \gets\operatorname{RMSNorm}(\mathbf{X}) $
\STATE $\mathbf{Q} \gets \widehat{\X}  \mathbf{W_q} + \bold{1}_nb_q^{\top}$ 
\STATE $\mathbf{P} \gets \widehat{\X} \mathbf{W_{kv}} + \bold{1}_nb_{kv}^{\top}$
\STATE \textbf{Compute multi-head importance scores}
\STATE $\mathbf{Z} \gets \operatorname{MultiHeadImportanceScore}(\mathbf{X})$
\STATE $\mathbf{Z} \gets \mathbf{Z} + \operatorname{Gumbel}(0, 1).\operatorname{sample()}$
\STATE \textbf{Concatenate with 2k support vector paddings}
\STATE $\mathbf{P} \gets \operatorname{Concatenate}(\mathbf{P}, \mathbf{P_{\text{supp}}})$
\STATE $\mathbf{Z} \gets \operatorname{Concatenate}(\mathbf{Z}, \mathbf{Z}_\text{supp})$
\STATE \textbf{Sample sets of key and value vectors}
\STATE $\tilde{\mathbf{P}} \gets \operatorname{MultiHeadSample}(\mathbf{Z}, \mathbf{P}, k)$
\STATE $\tilde{\mathbf{K}}, \tilde{\mathbf{V}} \gets \operatorname{Split}(\tilde{\mathbf{P}})$
\STATE \textbf{Attend and transform tokens}
\STATE $\mathbf{X} \gets \mathbf{X} + \operatorname{FlashAttention}(\mathbf{Q}, \tilde{\mathbf{K}}, \tilde{\mathbf{V}}) $
\STATE $\widehat{\X} \gets\operatorname{RMSNorm}(\mathbf{X}) $
\STATE $\mathbf{X} \gets \mathbf{X} + \operatorname{FFN}(\widehat{\X})$
\STATE \textbf{Return} $\mathbf{X}$
% \Return $\X$
\end{algorithmic}
\end{algorithm}

\subsection{Parallelizable Differentiable Top-k Sampling} \label{sec:samsa}
To be a viable key-value sampler for transformers, that method needs both great parallelizability and reliable gradients for efficient model training. Here, we introduce the two methods that hold the needed properties: trivial differentiable sampling with replacement and our proposed differentiable without replacement method.

\paragraph{\textbf{With-Replacement Sampling.}} We define $z$ as the importance score of vectors, representing the vectors' likeliness of being picked. The importance scores of vectors are computed via a neural network: $\operatorname{ImportanceScore}(\mathbf{X}): \mathbb{R}^{n \times m} \rightarrow \mathbb{R}^{n}$:
\begin{equation*} \label{eq:gumbel_distribution}
    z \triangleq \operatorname{ImportanceScore}(\bold{X}).
\end{equation*}
This simple method conditions the importance score of each token on the entire set of tokens, allowing patterns with complex dependencies to emerge. If we allow the set of sampled vectors to contain duplicates, Gumbel-Top-k algorithm can be parallelized simply by computing everything concurrently as follows:
\begin{align*}
    \operatorname{Sample}_1(z, \mathbf{X}, k) &\triangleq \operatorname{Concatenate}_{i=1}^k(\operatorname{ST-GumbelSoftmax}(z)^{\top}\mathbf{X}).
\end{align*}

\paragraph{\textbf{Without-Replacement Sampling.}} The weakness of sampling with replacement is that it creates duplicates. These duplicates waste computational resources on the same pieces of information, which harms the expressivity of transformers relying on sampling. This proposed method of without-replacement sampling relies on a novel proxy soft gradient estimator we devised. To begin with, we started with a hypothetical inefficient but parallelizable method: select one set of sampled vectors from sets of all sets of sampled vectors using Gumbel-Softmax (this is the definition in Equation~\ref{eq:naive_sampler}). This is equivalent to a continuous relaxation of the discrete brute-force combinatorial optimization algorithm. By defining the likeliness of choosing one set (i.e. the importance score of sets of vectors) as the sum of the importance scores of its elements, we have:
\begin{align*}
    z^* &\triangleq \operatorname{Concatenate}_{\mathbf{S}}^{C(\mathbb{N}^{\leq n},k)}(\sum_i^\mathbf{S} z_i), \\
    \mathbf{X}^* &\triangleq \operatorname{Concatenate}_{\mathbf{S}}^{C(\mathbb{N}^{\leq n},k)}(\operatorname{vec}(\mathbf{X}_\mathbf{S}) ^ {\top}), \\
    \operatorname{Sample}_2(z, \mathbf{X}, k) &\triangleq \operatorname{ST-GumbelSoftmax}(z^*)^{\top}\mathbf{X}^*,
\end{align*}
where $C(N,k)$ is the set of all k-combinations of set $N$, $\mathbb{N}^{\leq n}$ is the set of all natural numbers less than or equal to $n$, $\operatorname{vec}$ is the matrix vectorization function obtained by stacking the columns of the input matrix on top of one another, $\mathbf{X}_\mathbf{S}$ is the matrix of the row vectors with indices in $\mathbf{S}$ of $\mathbf{X}$. This definition of the importance scores of sets has a clear advantage: not needing to enumerate over the entire $C(n,k)$ at inference. Since using all possible combinations to generate one gradient vector for a single optimization step is overkill, an immediate improvement is local search - a sequential algorithm. Intuitively, this allows us to put the optimization steps of local search and gradient descent in the same for loop. For local search, we define the current solution as the set with the highest importance score. The locality of an index set $S$ is then described as the collection of index sets, each differing from $S$ by exactly one element:
\begin{align*}
    \operatorname{Current} &\triangleq \operatorname{argmax}(z^*) = \operatorname{ArgTopK}(z)\\
    \operatorname{Local}(\mathbf{S}) &\triangleq \{x: x \in C(\mathbf{X}, k) \text{ and } \left|x - \mathbf{S}\right| = 1\}
\end{align*}
Using local search as the base discrete combinatorial optimization algorithm, we compare the current solution against one of its localities, using Gumbel-Softmax reparameterization. The average result from comparing $k(n-k)$ pairs of solutions is given as follows:

\begin{align*}
    i &\triangleq \operatorname{ArgTopK}(z, k),\\
    j &\triangleq \operatorname{ArgTopK}(-z, n - k),\\
    p &\triangleq J_{k, n - k}, \\ 
    \partial p / \partial z &\triangleq \partial \left(\operatorname{ST-GumbelSigmoid}\left(z_{i}\bold{1}_{n-k}^{\top} - \bold{1}_kz_{j}^{\top}\right)\right) / \partial z, \\
    \operatorname{Sample}(z, \mathbf{X}, k)[m] &\triangleq \sum_{v=1}^{n-k} \frac{\bold{X}[i_m] * p[m, v] + \bold{X}[j_v] * (1 - p[m, v])}{n- k},\\
\end{align*}
where $J_{m,n}$ is all-ones matrix with $m$ rows and $n$ columns, $\bold{1}_k$ are the all-one vectors of length $k$. The Gumbel Sigmoid function is the Gumbel Softmax when the number of categories is equal to two (details in Appendix~\ref{sec:misc}), similar to the relationship between Sigmoid and Softmax functions. To further cut down the optimization cost, we select $i$ as top-k most important and $j$ as top-k second most important index vectors.

\subsection{Graph Modeling}
\label{sec:graphModeling}

The complexity of the graph data structure is needed for the claim of working on many data modalities of ours. Unfortunately, existing literature deemed self-attention alone insufficient for graph and they created very complicated methods: e.g. combination between Message Passing Neural Network (MPNN) and self-attention layers~\citep{GraphGPS,Graphormer,GraphTransformerNetwork}. This makes it hard to adapt to other data modalities. We identify the reason transformers need so many modifications just to be barely working on graphs is the inability to process both the nodes and edges information simultaneously. Therefore, we propose Graph-Bridge, a simpler approach treating both nodes and edges as tokens, which can be applied to any self-attention network without complicated constructions.

\paragraph{\textbf{Formulation.}} Given a graph \( G = (V, E) \) where \( V \) represents the set of nodes and \( E \) represents the set of edges, the objective is to learn a mapping \( \mathcal{F}: G \rightarrow \mathbb{R}^d \) that captures the structural properties of \( G \) in a \( d \)-dimensional latent space. Each node \( v_i \in V \) is associated with a feature vector \( \mathbf{x}_i \in \mathbb{R}^f \), and each edge \( e_{ij} = (v_i, v_j) \in E \) is associated with a feature vector \( \mathbf{e}_{ij} \in \mathbb{R}^g \).

\paragraph{\textbf{Positional Encoding.}} To integrate node and edge information, we introduce a novel graph positional encoding (PE) mechanism. Let \( \mathbf{p}_i \in \mathbb{R}^d \) denote the positional encoding of node \( v_i \). We define \( \mathbf{p}_i \) as a random vector sampled from a normal distribution:
\[
\mathbf{p}_i \sim \mathcal{N}(0, \operatorname{diag}(\sigma^2)) + \phi(x_i),
\]
where \( \sigma^2 \) is a learnable variance $d$-dimensional vector and $\phi: \mathbb{R}^f \rightarrow \mathbb{R}^d$ is a multi-layer perceptrons (MLPs) to inject node feature into positional encodings. For each edge \( e_{ij} \), its positional encoding \( \mathbf{p}_{ij} \) is computed as:
\[
\mathbf{p}_{ij} = \mathbf{p}_i - \mathbf{p}_j.
\]
While summation is usually used to combine positional encodings, in this case, we use subtraction for edge encoding to preserve the directional information of graph edges.

\paragraph{\textbf{Node and Edge Transformation.}} To map node and edge representations into a common latent space, we use two separate MLPs, \( \phi_V: \mathbb{R}^d \times \mathbb{R}^f \rightarrow \mathbb{R}^d \) and \( \phi_E: \mathbb{R}^d \times \mathbb{R}^g \rightarrow \mathbb{R}^d \), defined as:

\begin{align*}
    \mathbf{h}_i &= \phi_V(\mathbf{p}_i, \mathbf{x}_i), \\
    \mathbf{h}_{ij} &= \phi_E(\mathbf{p}_{ij}, \mathbf{e}_{ij}),
\end{align*}
where \( \mathbf{h}_i \) and \( \mathbf{h}_{ij} \) are the transformed representations of node \( v_i \) and edge \( e_{ij} \), respectively. These features are then processed by subsequent transformer layers, allowing complex relationships between nodes and edges to be recognized. 

\paragraph{\textbf{Compare to other Graph-PE.}} This method is simple and efficient to compute: requiring only random feature generation, gathering operation on the randomly generated node PE for edges, and MLP features. This is different to existing graph positional encodings like Random-Walk PE~\citep{RWPE} (hard to parallelize) and Laplacian PE~\citep{GraphGPS} (inherently computationally expensive). While being different, how this method works is surprisingly similar to random-walk PE: each self-attention propagates signals through edge tokens, i.e. we have merged the sequential process of random-walk and multi-layer neural network altogether.

\subsection{Integration to Self-Attention Mechanism}
\label{sec:intSA}
While a straightforward integration of our sampling method into the self-attention mechanism is functional, we can achieve significant improvements. The following sections outline our enhancements.

\paragraph{\textbf{Mixture of Attention Experts.}} While global attention mechanisms are highly parallelizable and efficient, they limit each token's local receptive field since all tokens attend to the same set of selected tokens. Each attention head operates independently, making self-attention outputs invariant to token permutations within each head. Inspired by MoE (Mixture of Experts)~\citep{expert_choose_token}, we propose a method where each attention head selects a distinct subset of sampled tokens. This process is parallelizable since the $h$-sampling processes are independent. Formally, the multi-head importance score $\mathbf{Z}$ is computed by:
\begin{equation*}
    \mathbf{Z} \triangleq \operatorname{MultiHeadImportanceScore}(\bold{X}).
\end{equation*}
And the multi-head sampling function $\operatorname{MultiHeadSample}$ as follows:
\begin{align*}
    \mathbf{\tilde{P}} &\triangleq \operatorname{MultiHeadSample}(\mathbf{Z}, \mathbf{P}, k) \\ &= \operatorname{Stack}_{i=1}^{h}(\operatorname{Sample}(\mathbf{Z}_{\dots,i}, \mathbf{P}, k)),
\end{align*}
where $\mathbf{Z}_{\dots,i}$ is the $i^\text{th}$ column vector of $\mathbf{Z}$ matrix. By incorporating this mixture of attention experts, our sampling transformer significantly enhances its local receptive field by a factor of $h$. For instance, a transformer with four attention heads and a 25\% sampling rate can learn a full self-attention pattern if necessary.

\paragraph{\textbf{Learnable Support Vector Paddings.}} Simply attending all query vectors to a set of sampled key and value vectors has a limitation: when the number of tokens is less than k (the number of sampled tokens), our algorithm fails. To address this issue, we introduce 2k learnable key and value vectors, each associated with learnable importance scores, which are sampled concurrently with input tokens. The process can be described mathematically as follows:

\begin{align*}
    \mathbf{P} &\gets \operatorname{Concatenate}(\mathbf{P}, \mathbf{P_{\text{supp}}}), \\
    \mathbf{Z} &\gets \operatorname{Concatenate}(\mathbf{Z}, \mathbf{Z}_\text{supp})
\end{align*}
where $\mathbf{P}: \mathbb{R}^{n \times 2d}$ is the key-value matrix, $\mathbf{Z}: \mathbb{R}^{n \times h}$ is the multi-head importance score matrix, $\mathbf{P_\text{supp}: \mathbb{R}^{2k \times 2d}}$ is the learnable support vector matrix, and $\mathbf{Z_\text{supp}: \mathbb{R}^{2n \times h}}$ is the learnable multi-head importance score matrix. These vectors and learned importance scores function is much similar to support vectors and Lagrange multipliers in traditional kernel methods, effectively introducing additional decision boundaries within the self-attention mechanism. Consequently, this not only enhances the learning capacity of transformers but also ensures stable training by consistently attending to a fixed number of tokens.

\paragraph{\textbf{Architecture Details.}} 
Our transformer architecture employs the Pre-RMSNorm formulation~\citep{prenorm,rmsnorm}, which has been demonstrated to accelerate convergence rates. For the residual branch, we utilize a zero-initialized learnable scalar multiplier~\citep{ReZero}, with a modification: the learnable scalar multiplier is shared across all network layers. The self-attention modules in our architecture are adapted to incorporate our sampling mechanism, as discussed in detail. The importance score neural network for each self-attention module is implemented as a simple multi-layer perceptron, as the latent representations produced by the preceding transformer stack already exhibit sufficient complexity and expressiveness. Finally, to align with recent advancements in language modeling, we integrate flash attention~\citep{Flash1, Flash2} into our architecture, which offers higher throughput and reduced memory usage. The full algorithm is summarized in Algorithm~\ref{alg:sampling_transformer}, which is in the overview section.

\subsection{Complexity Analysis}

For an input token $\X\in\mathbb{R}^{n\times d}$ with query and key weights $W_Q,W_K\in\mathbb{R}^{d\times d_a}$ and $k$ sampled tokens, the computational costs per 1-head transformer layer are:

\begin{itemize}
\item Linear transformations for $\Q,\K,\bold{V}$: $O(nd^2)$,
\item Calculating token importance scores: $O(nd)$,
\item Selecting top-k tokens: $O(n + k)$ forward, $O(n + k^2)$ backward,
\item Attention between $n$ queries and top-k keys: $O(nkd_h)$,
\item FFN network: $O(nd^2)$.
\end{itemize}

For large number of tokens ($n\gg d$), the asymptotic time and space complexity of SAMSA layer is $O(n)$.
\section{Experiments} \label{sec:experiments}
\subsection{Experiment Design}
Our SAMSA models are either using hard/soft sampling indicating whether the straight through estimator is used or not. We have trained \textbf{45 models} to measure the effectiveness of our method, both the performance and efficiency-wise of our model, against full self-attention (Flash Attention v2~\citep{Flash2}) and other specialized SOTA models on three data modalities (Sequence, Point Cloud, and Graph) on 4 datasets (Long Range Arena (all five), ModelNet40, ShapeNetPart, and Long Range Graph Benchmark (only Peptides)).
% The reason we chose these datasets to experiment on is simple:
% \begin{itemize}
%     \item Our model is a global attention model, therefore, we measure our model performance both on the known global-attention-only model's weakness and strength
%         \begin{itemize}
%             \item Global attention models are known to be weak against local structure and some datasets exhibit very strong locality characteristics: 3 LRA datasets~\citep{nangia-bowman-2018-listops,Pathfinder,CIFAR10} and 2 LRGB datasets~\citep{Peptideds} 
%             \item Global attention models are known to be good at long-range modeling and the tasks on these datasets require information on the global scale: 3 LRA datasets~\citep{Pathfinder,ImdbReview,ACLAnthologyNetworkCorpus}, 2 Point Cloud datasets \citep{Modelnet40,Shapenet}
%         \end{itemize}
%     \item These three data modalities exhibit minimal shared inductive bias, preventing the exploitation of their inherent structures and consequently compromising generalizability
% \end{itemize}
To minimize any possible bias when comparing between baselines, we use no data augmentation and identical hyperparameters across different tasks. This also implies we conducted little hyperparameter tuning and mostly focused on the architectural design.

\subsection{Performance-Efficiency Trade-offs}
\setlength{\tabcolsep}{1.2pt}
\begin{table*}[]
\begin{small}
\begin{center}
% \end{center}
\caption{Performance/Runtime of the full self-attention \citep{AttentionAllYouNeed} compared to SAMSA with three different sampling sizes and two sampling formulations. \textbf{\textcolor{OliveGreen}{Green}} and \textbf{\textcolor{RawSienna}{Red}} indicate SAMSA having better or worse statistics compared to full self-attention, respectively. The results of full self-attention in LRA are obtained from~\citep{MEGA}. It has better performance than what reported in the dataset paper~\citep{longrangearena}. The results of full self-attention in LRGB are obtained from the dataset paper~\citep{LongRangeGraphBenchmark}. The results of full self-attention on Shape Classification on ModelNet40 is run by ourselves, with the exact same hyperparameters as other SAMSA models. The results of PartSegmentation on ShapeNetPart is obtained from PointBERT~\citep{PointBERT}.} %continue
\label{tab:against_full_self_attention}

\begin{tabular}{@{}cllrccccccc@{}}
\toprule
\multicolumn{4}{c}{}                                                                         & \multicolumn{6}{c}{\textbf{SAMSA}}                                                                                                                                                                                                                                                            & \textbf{Full Self-Attention} \\ \midrule
\multirow{2}{*}{\textbf{Exp Settings}} & \multicolumn{3}{l}{\textbf{Sampling Size}}          & \multicolumn{2}{c}{128}                                                                       & \multicolumn{2}{c}{256}                                                                       & \multicolumn{2}{c}{512}                                                                       & \multirow{2}{*}{N/A}         \\ \cmidrule(lr){2-10}
                                       & \multicolumn{3}{l}{\textbf{Soft/Hard Sampling}}     & Soft                                          & Hard                                          & Soft                                          & Hard                                          & Soft                                          & Hard                                          &                              \\ \midrule
\multirow{5}{*}{\textbf{Sequence}}     & \textbf{ListOPS}      & L=2048 & Acc$^{\uparrow}$   & \textbf{\textcolor{OliveGreen}{41.53}}        & \textbf{\textcolor{OliveGreen}{41.13}}        & \textbf{\textcolor{OliveGreen}{41.89}}        & \textbf{\textcolor{OliveGreen}{40.57}}        & \textbf{\textcolor{OliveGreen}{41.33}}        & \textbf{\textcolor{OliveGreen}{40.88}}        & 37.11                        \\ \cmidrule(l){2-11} 
                                       & \textbf{Pathfinder}   & L=1024 & Acc$^{\uparrow}$   & \textbf{\textcolor{OliveGreen}{81.97}}        & \textbf{\textcolor{RawSienna}{71.83}}         & \textbf{\textcolor{OliveGreen}{80.55}}        & \textbf{\textcolor{RawSienna}{69.71}}         & \textbf{\textcolor{OliveGreen}{79.74}}        & \textbf{\textcolor{OliveGreen}{72.28}}        & 71.83                        \\ \cmidrule(l){2-11} 
                                       & \textbf{Image}        & L=1024 & Acc$^{\uparrow}$   & \textbf{\textcolor{OliveGreen}{48.64}}        & \textbf{\textcolor{OliveGreen}{48.24}}        & \textbf{\textcolor{OliveGreen}{48.73}}        & \textbf{\textcolor{OliveGreen}{47.38}}        & \textbf{\textcolor{OliveGreen}{48.24}}        & \textbf{\textcolor{OliveGreen}{46.75}}        & 42.94                        \\ \cmidrule(l){2-11} 
                                       & \textbf{Text}         & L=2048 & Acc$^{\uparrow}$   & \textbf{\textcolor{RawSienna}{64.74}}         & \textbf{\textcolor{OliveGreen}{65.42}}        & \textbf{\textcolor{OliveGreen}{65.53}}        & \textbf{\textcolor{OliveGreen}{65.42}}        & \textbf{\textcolor{OliveGreen}{65.38}}        & \textbf{\textcolor{OliveGreen}{65.42}}        & 65.21                        \\ \cmidrule(l){2-11} 
                                       & \textbf{Retrieval}    & L=4096 & Acc$^{\uparrow}$   & -                                             & \textbf{\textcolor{OliveGreen}{82.05}}        & -                                             & \textbf{\textcolor{OliveGreen}{81.89}}        & -                                             & \textbf{\textcolor{OliveGreen}{81.81}}        & 79.14                        \\ \midrule
\multirow{2}{*}{\textbf{Graph}}        & \textbf{Pept-func}    & L=450  & AP$^{\uparrow}$    & \textbf{\textcolor{OliveGreen}{72.01}}        & \textbf{\textcolor{OliveGreen}{72.01}}        & \textbf{\textcolor{OliveGreen}{71.42}}        & \textbf{\textcolor{OliveGreen}{72.21}}        & -                                             & \textbf{\textcolor{OliveGreen}{72.45}}        & 63.26                        \\ \cmidrule(l){2-11} 
                                       & \textbf{Pept-struct}  & L=450  & MAE$^{\downarrow}$ & \textbf{\textcolor{OliveGreen}{0.2519}}       & \textbf{\textcolor{OliveGreen}{0.2458}}       & -                                             & \textbf{\textcolor{OliveGreen}{0.2515}}       & -                                             & \textbf{\textcolor{OliveGreen}{0.2468}}       & 0.2529                       \\ \midrule
\multirow{2}{*}{\textbf{Point Cloud}}  & \textbf{ModelNet40}   & L=1024 & Acc$^{\uparrow}$   & \textbf{\textcolor{OliveGreen}{91.11}}        & \textbf{\textcolor{RawSienna}{90.91}}         & \textbf{\textcolor{OliveGreen}{91.72}}        & \textbf{\textcolor{RawSienna}{90.95}}         & \textbf{\textcolor{OliveGreen}{91.35}}        & \textbf{\textcolor{OliveGreen}{91.39}}        & 91.07                        \\ \cmidrule(l){2-11} 
                                       & \textbf{ShapeNetPart} & L=2500 & IoU$^{\uparrow}$   & -                                             & \textbf{\textcolor{OliveGreen}{85.74}}        & -                                             & \textbf{\textcolor{OliveGreen}{85.87}}        & -                                             & \textbf{\textcolor{OliveGreen}{85.62}}        & 85.1                         \\ \midrule
\multicolumn{4}{c}{\textbf{Inference Speed (1024 tokens)}}                                   & \textcolor{OliveGreen}{$\mathbf{1.63\times}$} & \textcolor{OliveGreen}{$\mathbf{2.17\times}$} & \textcolor{OliveGreen}{$\mathbf{1.44\times}$} & \textcolor{OliveGreen}{$\mathbf{1.62\times}$} & \textcolor{OliveGreen}{$\mathbf{1.08\times}$} & \textcolor{OliveGreen}{$\mathbf{1.44\times}$} & 1$\times$                    \\ \bottomrule
\end{tabular}
\end{center}
\end{small}
\end{table*}
We evaluated the performance of our SAMSA model on nine different datasets using six distinct SAMSA model configurations. The detailed results is presented in Table~\ref{tab:against_full_self_attention}.
\paragraph{\textbf{Trade-offs against Full Self-Attention.}} At a 50\% sampling rate, SAMSA is significantly faster and cheaper to compute than full self-attention, particularly due to the hard formulation, which avoids the need for gradient estimation during inference. Performance-wise, SAMSA consistently outperforms full self-attention, even at very low sampling rates (e.g., Retrieval: 128 / 4k). This superior performance is likely due to the regularization effect introduced by the sampling operator. Early in training, the random selection of tokens acts like DropAttention regularization~\citep{dropattention}, preventing overfitting by ensuring attention is not overly concentrated on specific tokens. As training progresses, the sampler becomes more effective at identifying important tokens, reducing the regularization effect and allowing for more targeted attention. This dynamic aligns with known phenomena where early-stage regularization accelerates convergence, leading to improved overall performance~\citep{dropunderfit}.

\paragraph{\textbf{Trade-offs between Different Sampling Rates.}} Efficiency-wise, Table~\ref{tab:against_full_self_attention} shows that SAMSA's inference speed decreases with lower sampling rates and can be twice as fast as full self-attention on an A100. Performance-wise, the table also shows that a sampling rate of 256 tokens per head outperforms both full self-attention and higher sampling rates, suggesting that not all tokens equally impact attention effectiveness. This "sampling nonlinearity" may help filter out redundant tokens. Additionally, SAMSA's performance on easy tasks like LRA-Text and LRA-Image decreases as sampling rates increase, indicating a potentially stronger regularization effect of lower sampling rates.

\paragraph{\textbf{Trade-offs between Hard and Soft Sampling.}} Hard-SAMSA is much faster than soft-SAMSA, with a 33\% speed advantage at a 50\% sampling rate. However, hard-SAMSA underperforms in certain tasks, likely due to discrepancies between the forward and backward passes in Gumbel-Softmax~\citep{GumbelSoftmax}. Notably, hard-SAMSA's learning curves in the Pathfinder task are volatile, reflecting the adverse effects of the straight-through estimator. Interestingly, earlier navigation through "breakpoints" leads to worse performance, possibly due to flat local minima, as shown by the stair-like learning curves (figure in Appendix~\ref{fig:curves}). However, it should be noted that hard-SAMSA performance is not inferior to soft-SAMSA with a counter-example: superior performance on graph data modality.

% \begin{figure}[]
% \begin{center}
% \includegraphics[width = 0.4\textwidth]{figure/pathfinder_curve.pdf}
% \caption{Learning curves (EMA=5) of SAMSA models with different configurations in LRA-Pathfinder task.}
% \label{fig:pathfinder_curve}
% \end{center}
% \end{figure}

\subsection{Sampling vs. Specialized Models}

Aside from comparing against full self-attention, we also compare our model against the SOTA-specialized model on each of our experimented data modalities (sequences, graph, and point cloud).

\paragraph{\textbf{Sequences.}} 
\setlength{\tabcolsep}{5pt}
\begin{table}[]
\begin{small}
\caption{(Long Range Arena) Accuracy on the full suite of long range arena (LRA) tasks. \textbf{Bold} indicates best sequence-models while \underline{underline} indicates best data modality agnostic models.}
% \vskip -0.1in
\label{tab-lra}
\begin{center}
% \begin{sc}

\begin{tabular}{@{}lcccccc@{}}
\toprule
\textbf{Method}                         & \textbf{ListOps} & \textbf{Text}   & \textbf{Retrieval} & \textbf{Image}  & \textbf{Pathfinder} & \textbf{Modality} \\ \midrule
BigBird~\citep{BigBird}                  & 36.05            & 64.02           & 59.29              & 40.83           & 74.87               & Sequence          \\
Reformer~\citep{kitaev2020reformer}      & 37.27            & 56.10           & 53.40              & 38.07           & 68.50               & Any               \\
Performer~\citep{performer}              & 18.01            & 65.40           & 53.82              & 42.77           & 77.05               & Any               \\
Linformer~\citep{wang2020linformer}      & 35.70            & 53.94           & 52.27              & 38.56           & 76.34               & Any               \\
Luna-256~\citep{ma2021luna}              & 37.98            & \underline{65.78} & 79.56              & 47.86           & 78.55               & Any               \\ \midrule
Transformer~\citep{AttentionAllYouNeed} & 37.11            & 65.21           & 79.14              & 42.44           & 71.83               & Any               \\ \midrule
S4~\citep{S4}                           & \textbf{88.65}   & 76.02           & 87.09              & 86.09           & 86.05               & Sequence          \\
MEGA~\citep{MEGA}                       & 63.14            & \textbf{90.43}  & \textbf{91.25}     & \textbf{90.44}  & \textbf{96.01}      & Sequence          \\ \midrule
hard-SAMSA (Ours)                       & 41.12            & 65.42           & 81.89              & 48.24           & 72.27               & Any               \\
soft-SAMSA (Ours)                       & \underline{41.88}  & 65.53           & \underline{82.05}    & \underline{48.73} & \underline{81.97}     & Any               \\ \bottomrule
\end{tabular}

\end{center}
\end{small}
\end{table}
In Table~\ref{tab-lra}, our model has competitive results against many other efficient transformers and the full attention transformer. From the table, it can be seen that MEGA~\citep{MEGA} and S4~\citep{S4} have the highest performance. This is true and expected because recurrent and locality via EMA are very strong inductive bias, making the model only work for sequences. Therefore, it is natural that our method, which is designed for multiple-data modalities, cannot outperform methods like those. Against weaker to no inductive bias like the rest, we achieved better performance on every task except text classification (losing Luna-256 by 0.25\%).

\paragraph{\textbf{Graph.}} 
\begin{table}[t]
\centering 
\small

\caption{Experimental results on Peptides func and Peptides struct. The results of methods other than ours are taken from their respective works and LRGB~\citep{LongRangeGraphBenchmark}. We report the means and standard deviations of 4 random seeds.}
\label{tab-lrgb}
\begin{tabular}{@{}l|cc@{}}
\toprule
\multicolumn{1}{l}{\textbf{Method}} &
\begin{tabular}[c]{@{}c@{}}\textbf{Peptides Struct}\\ \textbf{MAE} $^{\downarrow}$\end{tabular} & \begin{tabular}[c]{@{}c@{}}\textbf{Peptides Func}\\ \textbf{AP} $^{\uparrow}$ \end{tabular} \\ \midrule
GCN \citep{GCN} & 0.3496 ± 0.0013 & 0.5930 ± 0.0023 \\ 
GINE \citep{gin} & 0.6346 ± 0.0071 & 0.5498 ± 0.0079 \\
GatedGCN \citep{gated_gcn} & 0.3420 ± 0.0013 & 0.5864 ± 0.0077 \\
GatedGCN + RWSE \citep{gated_gcn} & 0.3357 ± 0.0006 & 0.6069 ± 0.0035  \\
\midrule 
Drew-GCN + LapPE \citep{drew} & 0.2536 ± 0.0015  & 0.7150 ± 0.0044  \\
GraphGPS + LapPE \citep{gps} & 0.2500 ± 0.0005 & 0.6535 ± 0.0041 \\ 
GRIT \citep{RWPE}         & {0.2460 ± 0.0012}  & 0.6988 ± 0.0082   \\

Graph VIT \citep{graphmlpmixer}       & \textbf{0.2449 ± 0.0016}  & 0.6942 ± 0.0075   \\
GraphMLP Mixer \citep{graphmlpmixer}  & 0.2475 ± 0.0015  & 0.6921 ± 0.0054    \\

MPNN + VN + RWSE \citep{cai2023connection} & 0.2529 ± 0.0009  & 0.6685 ± 0.0062    \\ 

\midrule

24-layer hard-SAMSA (ours)& \underline{0.2458 ± 0.0013} & \textbf{0.7221 ± 0.0042}\\ 
24-layer soft-SAMSA (ours)& 0.2519 ± 0.0011 & \underline{0.7171 ± 0.0051}\\ 
\bottomrule
\end{tabular}
\end{table}
In Table~\ref{tab-lrgb}, our model has outperformed most methods in the two tasks we tested. This shows the efficacy of our newly proposed Graph-PE. Note that, our method's learnable positional encoding is computed cheaply, and how cheap the computation is best represented in Table \ref{tab-lrgb-runtime}. The table shows that our entire model computes everything even before other methods finish their positional encoding computation. It also shows the efficacy of the SAMSA model as a very deep network, which is also different from traditional graph learning methods which cannot extend beyond 8 layers due to over-smoothing.

\begin{table}[]
\begin{small}
\begin{center}
\caption{Runtime of Graph Positional Encoding on one epoch on A100 compared to our 24-layer SAMSA model. The runtime of LapPE and RWPE are taken from \citep{LongRangeGraphBenchmark}.}
\label{tab-lrgb-runtime}
\begin{tabular}{@{}llll@{}}
\toprule
\textbf{Method}          & LapPE                   & RWPE                    & SAMSA (whole model)              \\ \midrule
\textbf{Peptides Struct} & \multicolumn{1}{c}{73s} & \multicolumn{1}{c}{53s} & \multicolumn{1}{c}{\textbf{35s}} \\ \bottomrule
\end{tabular}
\end{center}
\end{small}
\end{table}

\begin{table}[]
\begin{small}
\begin{center}
\caption{Inference time (in seconds) of the networks for ModelNet40 classification test split in 1 A100 and 8 CPUs with a batch size of 32. The results of PointNet and DGCNN are taken from~\citep{LearntEquivariance}.}
\label{tab-pointcloud-runtime}
\begin{tabular}{@{}lllll@{}}
\toprule
\textbf{Method}         & PointNet                & DGCNN                   & hard-SAMSA (ours)                & soft-SAMSA (ours)               \\ \midrule
\textbf{Inference Time} & \multicolumn{1}{c}{18s} & \multicolumn{1}{c}{23s} & \multicolumn{1}{c}{\textbf{<1s}} & \multicolumn{1}{c}{\textbf{1s}} \\ \bottomrule
\end{tabular}
\end{center}
\end{small}
\end{table}

\begin{table}[H]
\caption{Experimental results on ModelNet40 dataset~\citep{Modelnet40} and part segmentation results on ShapeNetPart dataset~\citep{Shapenet} against point-cloud specialized baselines.}
% \vskip -0.1in
\label{table:pointcloud}
            \centering
\begin{small}
% \begin{sc}
\begin{tabular}{l|cccc}
\toprule
\multirow{2}{*}{\textbf{Method}} & \multicolumn{2}{c}{\textbf{ModelNet40}}      & \multicolumn{2}{c}{\textbf{ShapeNetPart}}                \\ 
\cmidrule(r){2-5}
                        & \textbf{mAcc}${}^\uparrow$ & \textbf{OA}${}^\uparrow$ & \textbf{c. IoU}${}^\uparrow$ & \textbf{i. IoU}${}^\uparrow$ \\
\midrule
PointNet~\citep{PointNet}               & 86.2              & 89.2            & 80.4                   & 83.7                   \\
Set Transformer~\citep{SetTransformer}         & -                 & 90.4            & -                      & -                      \\
PointNet++~\citep{PointNet++}              & -                 & 91.9            & 81.9                   & 85.1                   \\
SpecGCN~\citep{wang2018local}                 & -                 & 92.1            & -                      & -                      \\
PointCNN~\citep{PointCNN}                & 88.1              & 92.2            & 84.6                   & 86.1                   \\
DGCNN~\citep{DynamicGraphCNN}                   & 90.2              & 92.2            & 82.3                   & 85.1                   \\
PointWeb~\citep{PointWeb}                & 89.4              & 92.3            & -                      & -                      \\
SpiderCNN~\citep{SpiderCNN}               & -                 & 92.4            & 81.7                   & 85.3                   \\
PointConv~\citep{PointConv}               & -                 & 92.5            & 82.8                   & 85.7                   \\
Point2Sequence~\citep{Point2Sequence}          & 90.4              & 92.6            & -                      & 85.2                   \\
KPConv~\citep{KPConv}                  & -                 & 92.9            & \textbf{85.1}                   & 86.4                   \\
InterpCNN~\citep{InterpolatedConvolutionalNetworks}               & -                 & 93.0            & 84.0                   & 86.3                   \\
Point Transformer~\citep{PointTransformer}       & \textbf{90.6}              & \textbf{93.7}            & 83.7                   & \textbf{86.6}                   \\
\midrule
hard-SAMSA (Ours)       & 88.9 & 91.3 & 83.8 & 85.8             \\
soft-SAMSA (Ours)       & 88.7 & 91.7 & - & -            \\
\bottomrule
\end{tabular}
% \end{sc}
\end{small}
% \vskip -0.2in
% \vspace{-0.cm}
\end{table}
\paragraph{\textbf{Point Cloud.}}
Performance-wise, Table~\ref{table:pointcloud} shows that SAMSA has competitive results against other point cloud models. This is mainly because our model does not include any data-modality-specific inductive bias (like the usage of kNN for locality heuristic); therefore, it is susceptible to overfitting in scenarios where data is not abundant. Our method is also much faster than others, up to 18$\times$ faster than PointNet (which is the most small, simple, efficient point cloud model), as shown in Table~\ref{tab-pointcloud-runtime}.
\section{Conclusion} \label{sec:conclusion}
SAMSA (SAMpling-based Self-Attention) mechanism addresses the limitations of traditional transformers by reducing the quadratic complexity typically associated with self-attention. Unlike other efficient transformers that are often data-modality-specific, SAMSA provides a versatile, context-aware linear complexity solution that can be applied across various data modalities. By employing a differentiable sampling method to focus on the most important tokens, SAMSA achieves competitive or even state-of-the-art results on multiple benchmarks. Additionally, SAMSA's sparse formulation during inference minimizes computational costs, offering faster inference times with little in performance, and in some cases, even outperforming full self-attention transformers.

\paragraph{\textbf{Limitations.}} Despite the promising results achieved with SAMSA, there are certain limitations on the misaligned gradients of hard sampling formulation, which makes our model hard to converge in more difficult optimization problems (Pathfinder). This misaligned gradient issue in hard sampling methods is mostly from the misestimation of gradients of the non-chosen path. We believe that future work should focus on more precise gradient estimation would result in an inference-fast model with much better accuracy.

\bibliography{tmlr}
\bibliographystyle{tmlr}

\appendix
\newpage
\appendix
\onecolumn

\section{Datasets}
\paragraph{\textbf{ModelNet40.}} The ModelNet40 dataset~\citep{Modelnet40} consists of 12,311 pre-aligned shapes divided into 40 classes, where the train and test sets consist of 9,843 instances and 2,468 instances respectively. ModelNet40 is the pioneer large-scale 3D CAD dataset. Unlike previous CAD datasets \citep{princetonshape}, ModelNet40 is the pioneer large-scale dataset that is diverse in terms of both class and samples per class.

\paragraph{\textbf{ShapeNetPart.}} The ShapeNetPart dataset consists of 16,881 pre-aligned 3D shapes from 16 categories and is a part of a larger dataset: ShapeNetCore (51,300 3D models)~\citep{Shapenet}. The 3D shapes in the ShapeNetPart dataset are annotated with 50 segmentation parts in total representing virtual real-world 3D semantic models. ShapeNet provides a diverse variety of shape annotations and corresponding shapes. The full ShapeNet dataset a is multitude containing upright and front orientation vectors, parts and keypoints, shape symmetries, but we only account for the part-segmenting task in our work.

\paragraph{\textbf{Long Range Arena.}} The Long Range Arena (LRA)~\citep{longrangearena} is a composition benchmark of 5 tasks: ListOps \citep{nangia-bowman-2018-listops}, ImDB review \citep{ImdbReview}, ACL Anthology Network \citep{ACLAnthologyNetworkCorpus}, Grayscaled CIFAR-10 \citep{CIFAR10}, and Pathfinder \citep{Pathfinder}. These five tasks are all classification and they feature very long sequences: ListOps (2,048 tokens), ImDB review (1,024 tokens), ACL Anthology Network (4,096 tokens), Grayscaled CIFAR-10 (1,024 tokens), and Pathfinder (1,024 tokens). All five tasks involves tackling long-ranged data structures and manifold categorizations, challenging networks's generalization powers as well as memory and time efficiencies.

\paragraph{\textbf{Long Range Graph Benchmark.}} The Long Range Graph Benchmark (LRGB)~\citep{LongRangeGraphBenchmark} is a graph composition of five datasets. But in our work, we only consider two peptide molecular tasks: Peptides-struct and Peptides-func. The chemical structures and features of these peptides are represented as graph of around 155 nodes and diameter of around 57. These challenges are suitable for evaluating long range receptive fields of graph-based deep learning networks. Peptides-struct and Peptides-func are much similar, except the former is a graph regression task while the other is a graph classification task.

\section{Training Strategies and Data Preprocessing}

We use the \verb|pytorch| framework~\citep{Pytorch} to conduct our empirical experiments.

Here is the list of the common hyperparameters:

\begin{itemize}
    \item Optimizer: AdamW~\citep{adamw}
    \item Batch size: 32
    \item Learning rate: 0.001
    \item $\beta_1, \beta_2$: 0.9, 0.999
    \item Learning scheduler: Cosine Annealing
    \item Learning rate warm-up: Linear Warmup-2000 steps
    \item Gradient Clipping: Max Gradient Norm (2.0)
\end{itemize}

Our training phase details can be visited in Table~\ref{imptraindetail}.

\setlength{\tabcolsep}{4pt}
\begin{table*}[t]
\caption{Training hyperparameter details.}
\label{imptraindetail}
\begin{center}
% \begin{sc}
\begin{tabular}{@{}llcccc@{}}
\toprule
\textbf{Task}                & \textbf{Dataset} & \textbf{Learning Rate} & \textbf{Gradient Clipping} & \multicolumn{1}{l}{\textbf{Weight Decay}} & \textbf{No. of Epochs} \\ \midrule
\multirow{2}{*}{Point Cloud} & ModelNet40       & 0.002                  & 2.0                        & 0.1                                       & 200/600                \\
                             & ShapeNetPart     & 0.002                  & 2.0                        & 0.1                                       & 600                    \\ \midrule
\multirow{5}{*}{Sequence}    & ListOPS          & 0.001                  & 2.0                        & 0.01                                      & 50                     \\
                             & Text             & 0.001                  & 2.0                        & 0.01                                      & 50                     \\
                             & Retrieval        & 0.001                  & 2.0                        & 0.01                                      & 50                     \\
                             & Image            & 0.001                  & 2.0                        & 0.01                                      & 50                     \\
                             & Pathfinder       & 0.001                  & 2.0                        & 0.01                                      & 50                     \\ \midrule
\multirow{2}{*}{Graph}       & Peptides-struct  & 0.001                  & 2.0                        & 0.01                                      & 240                    \\
                             & Peptides-func    & 0.0004                 & 2.0                        & 0.01                                      & 160                    \\ \bottomrule
\end{tabular}
% \end{sc}
\end{center}
\end{table*}

% \begin{table}[h]
%     \centering
%     \begin{tabular}{l|l|ccc}
%           \toprule
%         \multicolumn{2}{l|}{\textbf{Task/Dataset}} & Batch Size & No. of Epochs & Gradient Clipping \\ 
%         \midrule
%         \multirow{2}{*}{Relational Positional Modeling}  &  ModelNet40     & 64& 1000 &\multirow{2}{*}{None} \\
%         &  ShapeNetPart     & 32& 1000 &\\
%         \hline
%         \multirow{5}{*}{Long Range Interaction Modeling}&  ListOps    & 64 & 50 & \multirow{5}{*}{Clipping gradient norm (max 1.0)}\\
%         \hline
%           &  Text    & 32 & 50 &\\
%           &  Retrieval    & 32 & 50 &\\
%           &  Image    & 64 & 200 &\\
%           &  Pathfinder    & 64 & 200 &\\
%         \hline
%         \multirow{2}{*}{Non Sequential Modeling}  &  ModelNet40     & 64& 600 &\multirow{2}{*}{None} \\
%         &  ShapeNetPart     & 32& 300 &\\
%     \end{tabular}
%     \caption{Training strategy details.}
%     \label{imptraindetail}
% \end{table}

As for the input preprocessing, our works are as follows:
\begin{itemize}
    \item \textbf{Long Range Arena:} We simply use one-hot encoding vectors as inputs.
    \item \textbf{Long Range Graph Benchmark:} We input the whole graph, with both node and edge as tokens.
    \item \textbf{ModelNet40:} Following previous work~\citep{PointNet}, we use farthest point sampling to sample 1024 initial points with normals.
    \item \textbf{ShapeNetPart:} We operate on the initial 2,500 point-normal without any change to the dataset.
    
\end{itemize}

All of the classification task training processes utilize Cross Entropy Loss, without label smoothing normalization. For Peptides-struct, we use $\operatorname{SmoothL1Loss}$ for optimization as the measuring metric is Mean Absolute Error (MAE). For Peptides-Func, we use threshold = 0.5 for Average Precision computation, implying we do not tune threshold parameter based on training data, aligning with previous methods.

\section{Architectural Specification} \label{sec:arch_spec}
Table~\ref{impmodeldetail} shows our model's architectural hyperparameters, for reproducibility purposes.

\begin{table*}[h]
\caption{Architectural details.}
\label{impmodeldetail}
\begin{center}
\begin{tabular}{@{}lcccccc@{}}
\toprule
\textbf{Dataset}     & $\mathbf{n_{\textbf{depth}}}$ & $\mathbf{d_{\textbf{model}}}$ & $\mathbf{d_{\textbf{FFN}}}$ & $\mathbf{n_{\textbf{heads}}}$ & $\mathbf{p_{\textbf{dropout}}}$ & $\mathbf{p_{\textbf{droppath}}}$ \\ \midrule
ModelNet40           & 6                             & 256                           & 768                         & 8                             & 0.1                             & 0.1                              \\
ShapeNetPart         & 8                             & 256                           & 768                         & 8                             & 0.1                             & 0.1                              \\
LRA-ListOPS          & 6                             & 256                           & 768                         & 8                             & 0.1                             & 0.0                              \\
LRA-Pathfinder       & 6                             & 128                           & 512                         & 8                             & 0.3                             & 0.0                              \\
LRA-Text             & 4                             & 128                           & 512                         & 8                             & 0.3                             & 0.0                              \\
LRA-Image            & 8                             & 128                           & 512                         & 8                             & 0.3                             & 0.0                              \\
LRA-Retrieval        & 6                             & 128                           & 512                         & 8                             & 0.3                             & 0.0                              \\
LRGB-Peptides-Func   & 24                            & 128                           & 512                         & 8                             & 0.3                             & 0.2                              \\
LRGB-Peptides-Struct & 24                            & 128                           & 512                         & 8                             & 0.3                             & 0.2                              \\ \bottomrule
\end{tabular}
% \end{sc}
\end{center}
\end{table*}

LRA Tasks use Sinusoid-1D positional encodings.

\section{Miscellaneous} \label{sec:misc}

ST-Gumbel-Sigmoid's Formula:

\begin{align*}
    \operatorname{GumbelSigmoid}(\pi) &\triangleq \frac{1}{1 + \exp\left((\pi_j + g_j) / \tau\right)}, \\
    \operatorname{ST-GumbelSigmoid}(\pi) &\triangleq 1 
        \begin{cases}
            1,& \text{if } \pi \leq 0, \\
            0,& \text{otherwise},
        \end{cases} \\
    \partial{\operatorname{ST-GumbelSigmoid}} / \partial{\pi} &\triangleq \partial{\operatorname{GumbelSigmoid}} / \partial{\pi},
\end{align*}
where \( g_i \) are independent and identically samples from \text{Gumbel}(0,1) distribution, \( \pi_i \) are the unnormalized logits, and \( \tau \) is the temperature parameter that controls the smoothness of the distribution.

\section{Learning Curves of Our Models} \label{fig:curves}
\begin{figure}[]
    \centering
    \subfigure[]{\includegraphics[width=0.4\textwidth]{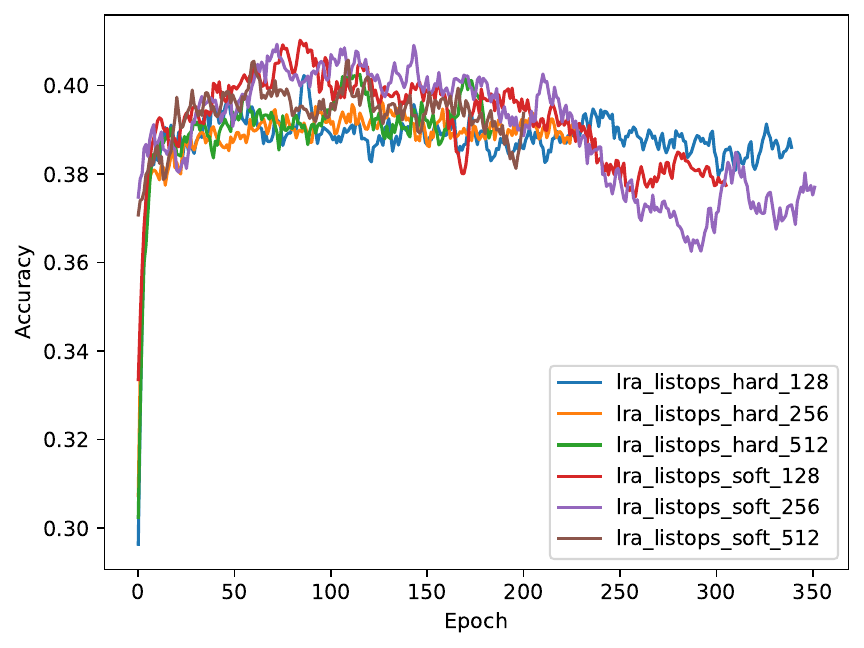}} 
    \subfigure[]{\includegraphics[width=0.4\textwidth]{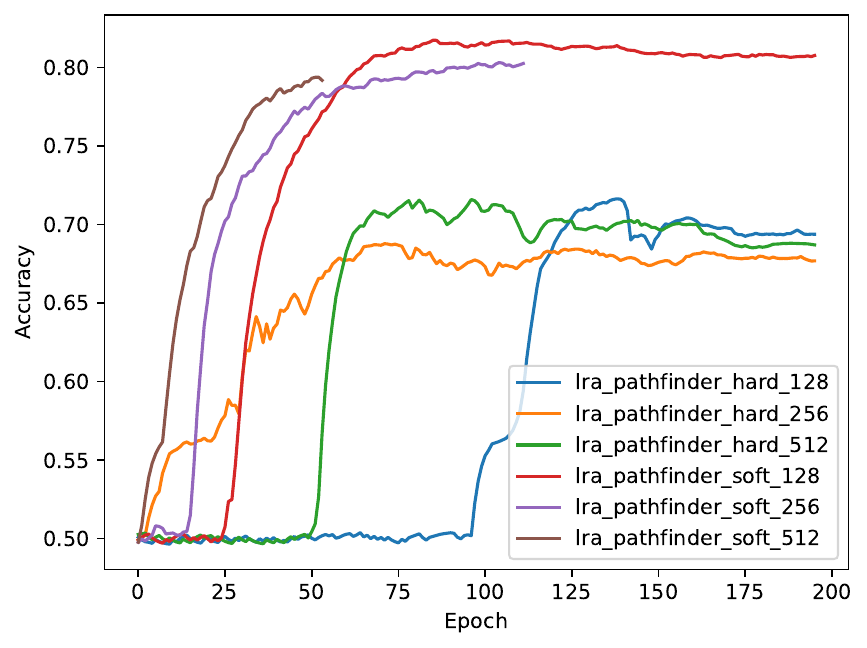}}
    \subfigure[]{\includegraphics[width=0.4\textwidth]{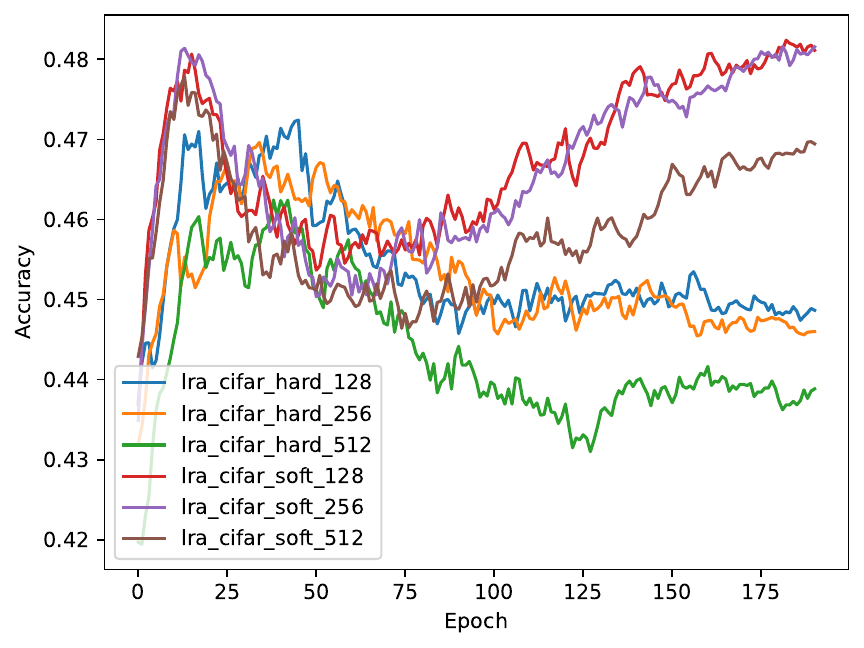}}
    \subfigure[]{\includegraphics[width=0.4\textwidth]{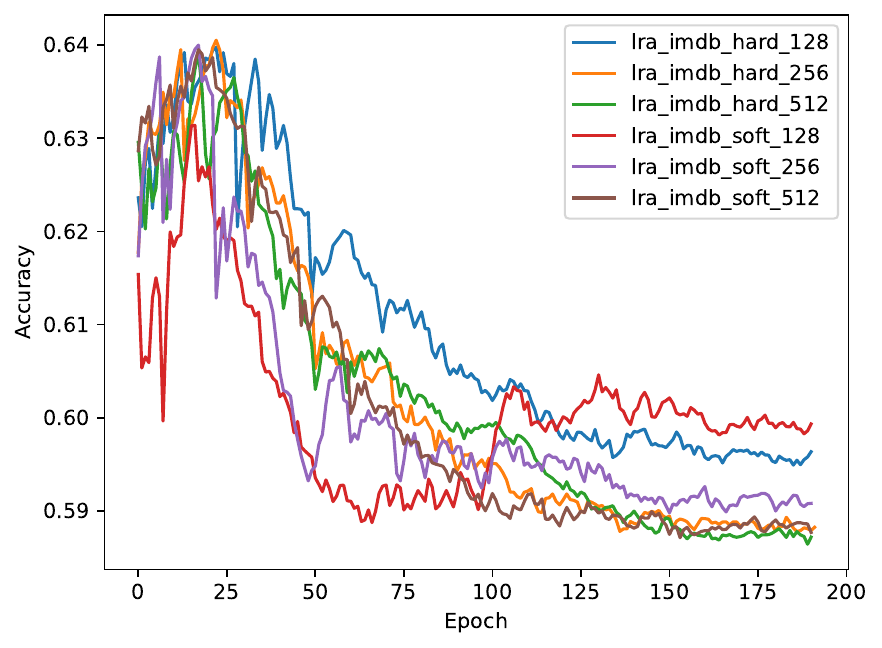}}
    \subfigure[]{\includegraphics[width=0.4\textwidth]{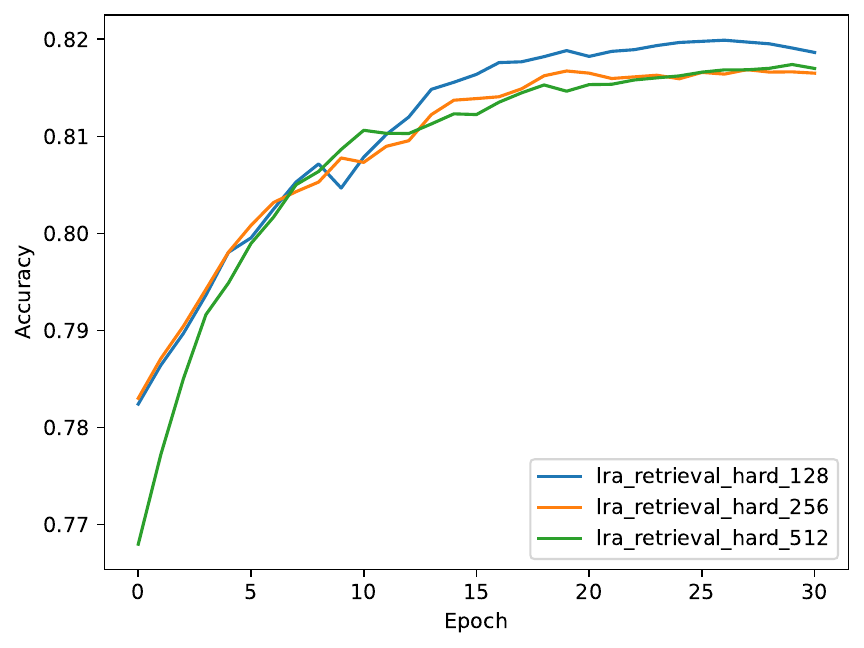}}
    \caption{Learning Curves of SAMSA models in Sequence Tasks}
    \label{fig:foobar}
\end{figure}

\begin{figure}[]
    \centering
    \subfigure[]{\includegraphics[width=0.4\textwidth]{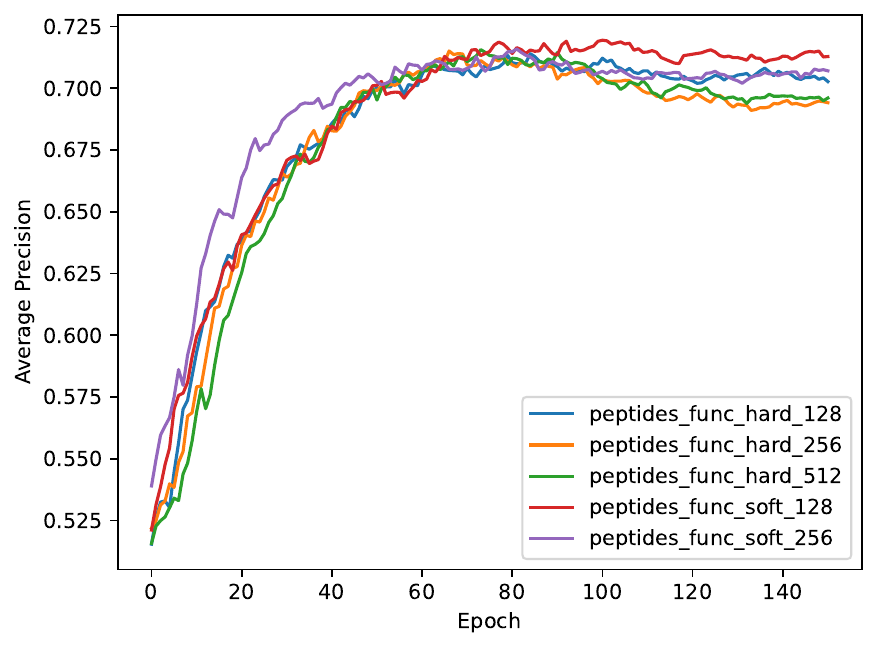}}
    \subfigure[]{\includegraphics[width=0.4\textwidth]{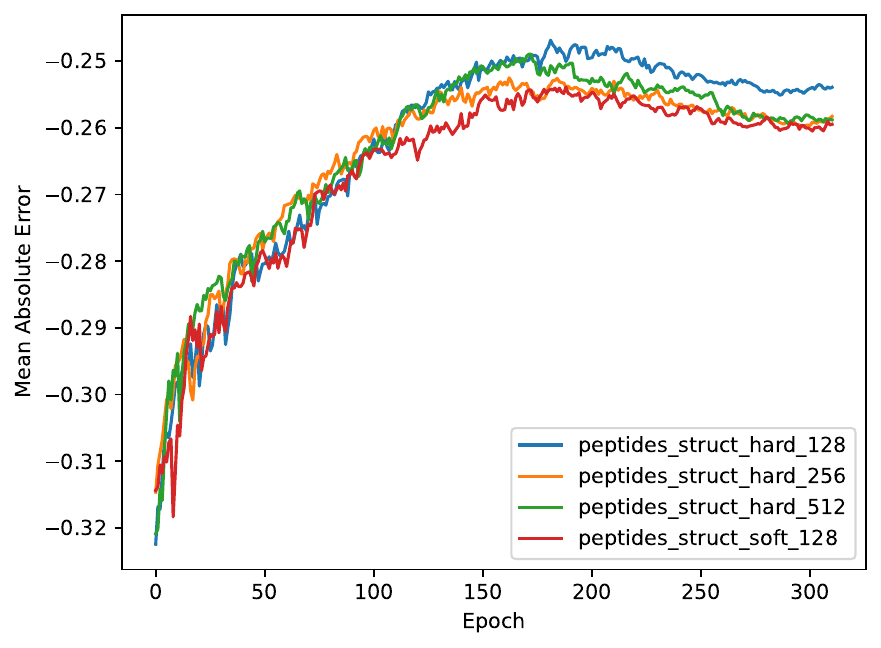}}
    \caption{Learning Curves of SAMSA models in Graph Tasks}
    \label{fig:foobar2}
\end{figure}

\begin{figure}[]
    \centering
    \subfigure[]{\includegraphics[width=0.4\textwidth]{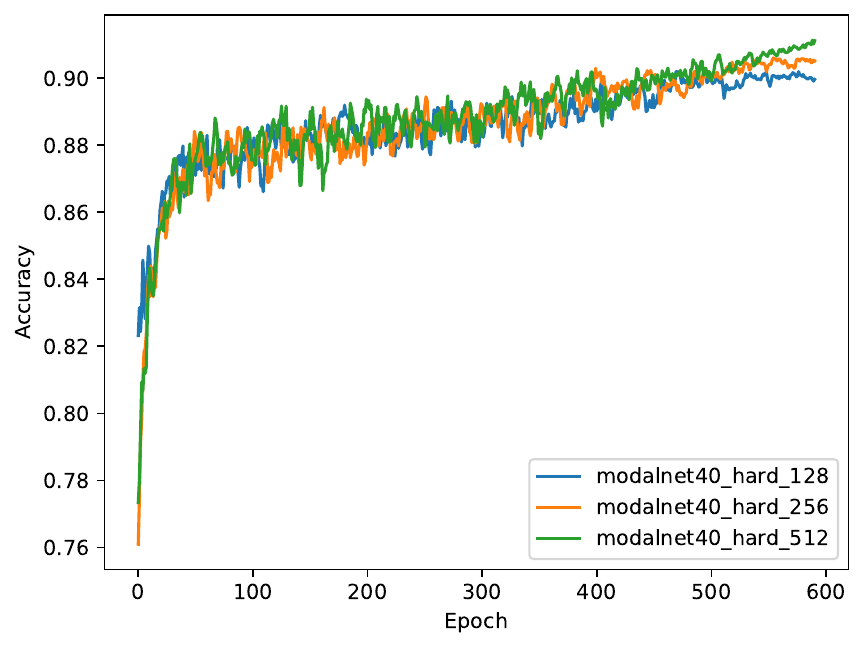}}
    \subfigure[]{\includegraphics[width=0.4\textwidth]{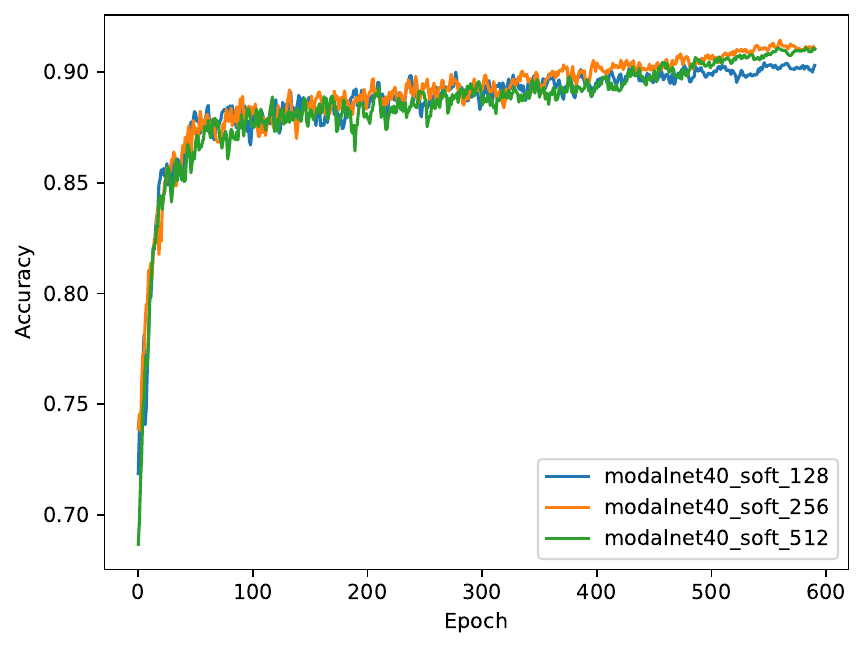}}
    \subfigure[]{\includegraphics[width=0.4\textwidth]{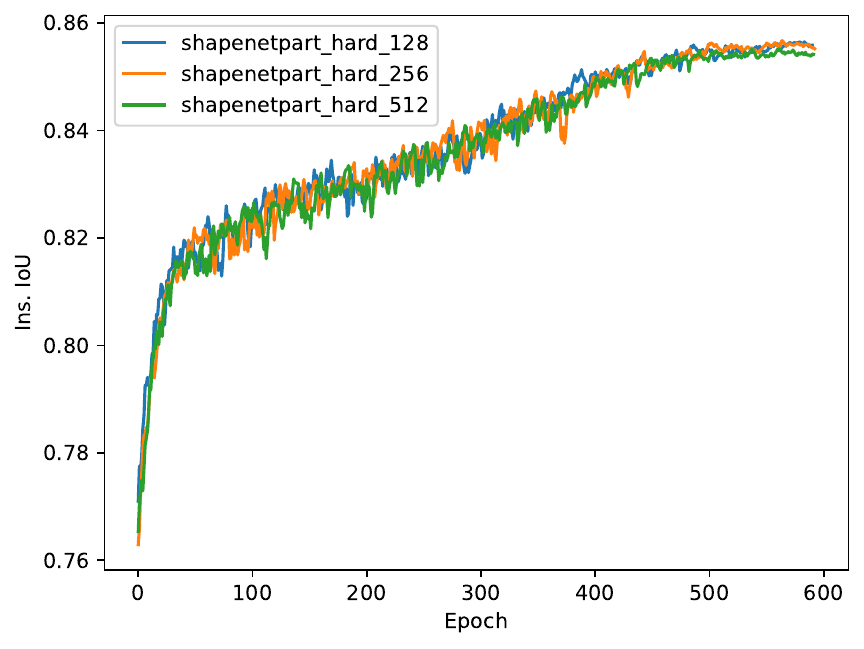}}
    \caption{Learning Curves of SAMSA models in Point Cloud Tasks}
    \label{fig:foobar3}
\end{figure}

Figure~\ref{fig:foobar}, ~\ref{fig:foobar2}, ~\ref{fig:foobar3} show the learning curve of SAMSA in various settings, providing for future understanding of how our model learned. All of the lines are plotted with EMA-5 smoothing.

\end{document}